%% file: root.tex
\documentclass[letterpaper, 10 pt, conference]{ieeeconf}  

\IEEEoverridecommandlockouts                              

\usepackage{times}
\usepackage{amsmath,amsfonts}
\usepackage{algorithmicx}
\usepackage{algorithm}
\usepackage{array}
\usepackage{textcomp}
\usepackage{stfloats}
\usepackage{url}
\usepackage{verbatim}
\usepackage{graphicx} 
\usepackage{xcolor}
\usepackage{threeparttable}
\usepackage{multirow}
\usepackage{mathrsfs}
\usepackage{amssymb}
\usepackage{algpseudocode}
\usepackage{url}
\usepackage{hyperref}
\usepackage{makecell} 
\usepackage{amssymb}  
\usepackage{algorithmicx}
\usepackage{algpseudocode}
\usepackage{subfigure}
\usepackage{multicol}
\usepackage{amsmath}
\usepackage{amsbsy}
\usepackage{amsfonts}
\usepackage{graphicx}
\usepackage{booktabs}    
\usepackage{multirow}    
\usepackage{array}       
\usepackage{caption}     
\usepackage{color}
\newcommand{\vpara}[1]{\vspace{1.5ex}\noindent\textbf{#1}}
\usepackage{listings}
\usepackage{tcolorbox}
\tcbuselibrary{listings, breakable, skins}

\definecolor{promptbg}{RGB}{245,245,245} 

\newtcblisting{promptbox}[1][]{
  enhanced,
  colback=promptbg,
  colframe=gray!80,          
  fonttitle=\bfseries\sffamily,
  title={#1},                 
  listing only,               
  listing options={
    basicstyle=\ttfamily\scriptsize, 
    breaklines=true,
    breakatwhitespace=true, 
    columns=flexible,       
    keepspaces=true,
  },
  top=0.5mm, bottom=0.5mm, left=2mm, right=2mm,
  boxrule=0.5pt,
  width=\textwidth 
}

\usepackage[export]{adjustbox}
\title{\LARGE \bf
Enabling a Unified Cross-Domain Representation for Two-Finger Gripper Manipulation via Interaction-Centric Modeling}

\author{Guanlin Li$^{1,\dagger}$, Shifeng Bao$^{1,\dagger}$, Yihan Zhao$^{1,\dagger}$, Haitao Shen$^{1,\dagger}$, Haoyang Li$^{1,\dagger}$, Chen Zhao$^{1,\dagger}$, \\Tong Yang$^{2,3}$, Jie Tang$^{2,3}$, and Jing Zhang$^{1}$%
\thanks{$^{\dagger}$Done as intern at Zhipu AI.}%
\thanks{$^{1}$Key Laboratory of Data Engineering and Knowledge Engineering, MOE, and School of Information, Renmin University of China, China.}%
\thanks{$^{2}$Zhipu AI, China.}%
\thanks{$^{3}$Department of Computer Science and Technology, Tsinghua University, China.}%
\thanks{Corresponding author: Jing Zhang ({\tt\small zhang-jing@ruc.edu.cn}).}%
}

\begin{document}

\maketitle
\thispagestyle{empty}
\pagestyle{empty}


\begin{figure*}[ht]
  \centering
  \includegraphics[width=\textwidth]{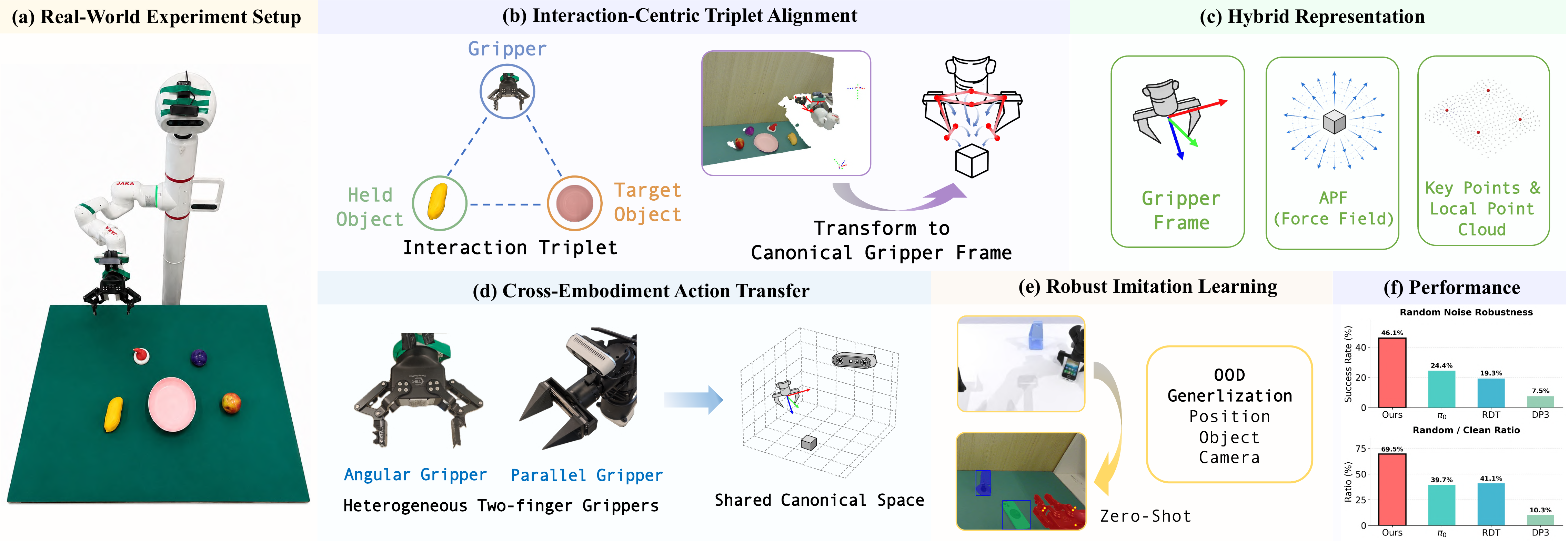}
  \caption{\textbf{Interaction-centric policy for robust two-finger manipulation.} We study universal two-finger manipulation on a real-world Jaka platform. By transforming RGB-D observations into a canonical gripper frame, our policy focuses on the interaction geometry among the gripper, held object, and target object. This formulation achieves stronger OOD generalization.}
  \label{fig:ICP}
\end{figure*}

\begin{abstract}

Achieving robust cross-embodiment generalization in imitation learning demands overcoming a critical representation flaw that inextricably entangles task semantics with hardware-specific visual geometry.
We propose an interaction-centric framework that leverages the shared structure of two-finger grippers via a parameterized universal gripper abstraction, yielding a canonical gripper-frame representation.
Given language and RGB-D observations, a VLM infers the subtask and grounds an interaction triplet (gripper, held, target), while SAM~2.1 tracks masks to reduce VLM queries.
We design concise hybrid features that combine target/collision artificial potential fields for global guidance with segmented gripper-frame point clouds for local geometry, and use a Flow-Matching Transformer to predict smooth 7-DoF action chunks.
Experiments in simulation and real-world tasks demonstrate that ours is the first imitation learning approach to simultaneously achieve competitive benchmark scores and extreme cross-embodiment/cross-viewpoint zero-shot sim-to-real transfer to completely distinct, heterogeneous robot platforms.

\end{abstract}

\input{latex/1.introduction}

\input{latex/2.related}

\input{latex/3.method}

\input{latex/4.experiment}

\input{latex/5.conclusion}


\bibliographystyle{ieeetr} 
\bibliography{reference}


\end{document}

%% file: latex/1.introduction.tex
\section{Introduction}
\label{sec:Introduction}

Imitation learning trains policies by regressing expert actions from environmental states~\cite{torabi2018behavioral, aloha, act}. However, in vision-centric manipulation, the robot embodiment becomes inherently coupled with visual observations, leading to severe overfitting to specific hardware setups~\cite{ rt1, rt2, openvla, hpt, x-vla, dp}. To achieve cross-embodiment transfer~\cite{umi, octo}, we hypothesize that a primary bottleneck lies in the embodiment representation itself. Successful transfer necessitates a structured representation that bridges perception and motor execution, moving beyond the paradigm of treating the robot merely as a passive visual feature.

Standard imitation learning seeks to learn a direct mapping from environmental observations to actions. To facilitate cross-domain generalization, we propose decoupling this end-to-end mapping into a two-stage process: projecting heterogeneous observations into a homogeneous representation space, and subsequently decoding actions from this intermediate space. In the context of manipulation, observations inherently comprise both visual and linguistic modalities. Therefore, this intermediate representation must serve as a semantic-geometric junction that fuses linguistic understanding with perceptual inputs. To this end, we identify \emph{interaction} as the essential structural decomposition of manipulation tasks. By distilling an interaction-centric representation from raw multimodal observations, we establish a unified, embodiment-agnostic interface from which the action sequence is ultimately generated.

We propose an appearance-agnostic, parameterized abstraction for two-finger grippers. Rather than relying on raw visual appearances, we extract the intrinsic geometric parameters of the gripper. This explicitly decouples the robot's kinematic structure from its visual appearance. By projecting heterogeneous robots into a unified canonical space based on these parameters, we establish the foundation for cross-embodiment generalization.

We ground this abstraction in intrinsic dimensions $[length, width, height]$ and reframe the manipulation task as reasoning over an \emph{interaction triplet}: the active gripper, any currently held, and the subsequent target (Fig.~\ref{fig:ICP}). To represent this triplet in the canonical frame, we define a parameterized \emph{keypoint envelope} that captures the fingertip contact points and collision boundaries with the opposite finger. Projecting observations into this coordinate system inherently aligns the underlying interaction geometries, substantially mitigating embodiment-specific variations.

Leveraging this abstraction, we propose an \emph{interaction-centric} pipeline (Fig.~\ref{fig:method}). A perception module uses a VLM to parse instructions and initialize SAM~2.1~\cite{sam} for real-time triplet tracking. A representation module lifts segmented regions to 3D, generating complementary features: Artificial Potential Fields (APFs)~\cite{khatib1986real} for coarse global guidance, and gripper-frame Point Clouds (PCDs) for fine-grained local contact geometry. Finally, a Flow-Matching Transformer~\cite{transformer,flowmatching} fuses these with semantics and proprioception to output 7-DoF trajectories, enabling diverse skills (e.g., pushing, rotating) beyond simple pick-and-place.

Our main contributions are summarized as follows:
\begin{itemize}
    \item \textbf{A unified interaction-centric paradigm.} We prove that interaction-centric representations unify cross-embodiment IL training, establishing a new paradigm for cross-embodiment pre-training.
    
    \item \textbf{A hybrid representation with strong generalization.} We propose a hybrid representation and IL system that achieves competitive end-to-end performance on benchmarks. This demonstrates that coarse encoding effectively aids IL generalization without sacrificing accuracy.
    
    \item \textbf{Data- and parameter-efficient zero-shot transfer.} Using only 50 demonstrations and 30M parameters, our lightweight model matches the in-domain performance of end-to-end baselines in simulation while maintaining strong generalization. Furthermore, we are the first to achieve simultaneous zero-shot cross-embodiment and cross-viewpoint transfer under compounded perturbations, spanning the sim-to-real gap, dual-arm to single-arm transfer, camera shifts, and object variations.    
\end{itemize}

%% file: latex/2.related.tex
\section{Related Work}
\label{sec:Related}


We situate our interaction-centric representation within the landscape of prior object-centric and keypoint-based paradigms, demonstrating how our parameterized gripper geometry and interaction triplets fundamentally overcome their cross-embodiment bottlenecks.

\subsection{Object-Centric Representation Learning}

Object-centric approaches~\cite{devin2018deep, controlvla, wu2025afforddp, heravi2023visuomotor} typically leverage visual models for object detection to focus on targets at the control level for success rates and generalization. Recently, several works in robotic grasping have shifted from object-centric to interaction-centric paradigms~\cite{zeng2024learning, wei2024dro, li2026mask2iv} to represent the interaction process itself. Unlike UMI~\cite{umi}, which captures interactions at the physical interface level, our work addresses interaction-centric paradigms at the representation level. Building upon this, we define interaction triplets, extend the interaction-centric representation to broader manipulation tasks, and implement a complete imitation learning system.


\subsection{Keypoint-Based Representation Learning}

Semantic object-keypoint representations~\cite{gao2023k, li2024unidoormanip,huangcopa} have significantly improved task generalization in robotic manipulation. Meanwhile, end-effector keypoints~\cite{fang2023anygrasp, chen2022keypoint, haldar2025point} act as intermediate representations that generalize across different robot shapes, which makes policy learning much more data-efficient. Recent studies have combined these two keypoint types~\cite{wang2025skil, huang2024rekep, pan2025omnimanip}, extending them to cross-embodied imitation learning and automatic generation, while integrating them with Vision-Language Models for stronger task abstraction. Building on this, we further decouple gripper keypoints from direct 3D object representations. By eliminating cumbersome object annotations, our approach achieves both data efficiency and cross-embodied generalization within a minimalist input space.


%% file: latex/3.method.tex
\section{Preliminaries}
We study language-conditioned manipulation with $N\in\{1,2\}$ robot arms, each equipped with a two-finger gripper.
At each time step $t$, a fixed and calibrated RGB-D camera provides an RGB image $I_t$ and a depth map $D_t$.

For gripper $i$, we denote its morphology by $\theta_i=[length_i,width_i,height_i]$ and its proprioceptive state by $\xi_{i,t}$ (the 7-DoF end-effector pose).
Let $\mathcal{T}$ denote known calibration parameters and geometric transformations (camera intrinsics/extrinsics and robot--world calibration) used to map depth observations into the gripper coordinate frame.
We define a gripper coordinate frame $\mathcal{G}_i$ whose origin is the full-closure fingertip reference point and whose axes align with the gripper orientation.

\paragraph{Observations}
Given a natural-language instruction $\mathcal{L}$, the observation for each gripper at time $t$ is
\begin{equation}
\mathbf{o}_{i,t} \triangleq \big(I_t, D_t, \mathcal{L}, \theta_i, \xi_{i,t}\big).
\end{equation}

\paragraph{Policy and actions}
Our goal is to learn an imitation policy $\pi_{\mathcal{T}}$, parameterized by $\mathcal{T}$, that operates at the \emph{perception time step} $t$ and outputs a horizon-$H$ action chunk executed at a finer control rate.
Formally,
\begin{equation}
\mathbf{A}_{i,t} = \pi_{\mathcal{T}}\!\left(\mathbf{o}_{i,t}\right),
\qquad
\mathbf{A}_{i,t} \triangleq [\mathbf{a}_{i,t,0}, \mathbf{a}_{i,t,1}, \dots, \mathbf{a}_{i,t,H-1}],
\end{equation}
where $\mathbf{a}_{i,t,h}\in\mathbb{R}^7$ denotes the 7-DoF gripper action at the $h$-th control step within perception step $t$, and $H$ is the action-chunk length.
$\mathbf{a}_{i,t,h}$ is represented as an end-effector pose increment plus a gripper open/close command, enabling frame changes via standard rigid-body adjoint transforms.

\begin{figure*}[t]
  \centering
  \includegraphics[width=\textwidth]{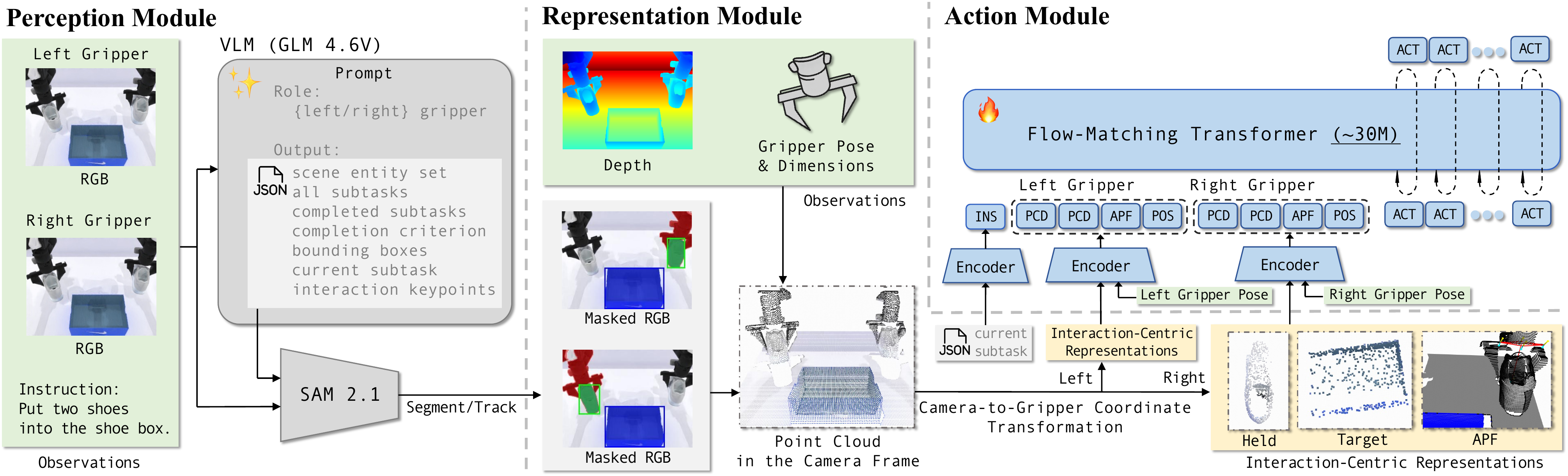} 
\caption{\textbf{Method overview.} The \textbf{Perception Module} utilizes a VLM for instruction parsing and SAM~2.1 for real-time triplet mask tracking. The \textbf{Representation Module} lifts 2D masks to 3D to produce disentangled artificial potential fields (APFs) and gripper-frame point clouds (PCDs). The \textbf{Action Module} employs a Flow-Matching Transformer to integrate subtask semantics, spatial features, and proprioception, generating smooth horizon-$H$ 7-DoF action trajectories.}
 \label{fig:method} 
\end{figure*}

\section{Method}
\label{sec:Method}

We propose to model the entire task through an interaction-centric formulation. At each timestep, we define an \textbf{interaction triplet}—the active gripper, the held object (if any), and the target object—while also constructing a highly compressed representation of the entire environment in the gripper frame. By combining the structured triplet with this compact global context, we preserve essential task-agnostic information, enabling robust action chunk prediction.

\subsection{Perception Module}
\label{sec:perception}

The Perception Module takes the instruction $\mathcal{L}$ and the current RGB observation $I_t$ and outputs the interaction-triplet masks at time $t$, shown in Algorithm~\ref{alg:perception}.
We use GLM-4.6V~\cite{hong2025glm} for instruction parsing and region localization, and SAM~2.1~\cite{sam} for mask generation and tracking.

\vpara{VLM task parsing and triplet localization.}
At VLM refresh steps ($t \bmod \delta~\footnote{$\delta$ is set to $5$ in our implementation.} = 0$), given $I_t$ and $\mathcal{L}$, we query the VLM (with gripper-$i$ prompts built from $\mathcal{L}$) to produce the current structured task context $\mathcal{C}_t$:
\begin{equation}
\mathcal{C}_t \leftarrow \mathrm{VLM}\!\left(I_t, \mathcal{L}\right),
\qquad
\mathcal{C}_t = \{\mathcal{C}_{i,t}\}_{i=1}^{N}.
\end{equation}

\vpara{Segmentation and tracking with SAM~2.1.}
To refine the VLM's coarse bounding boxes into pixel-accurate masks, we employ SAM~2.1 in two modes: at refresh steps, it segments $I_t$ using boxes from $\mathcal{C}_{i,t}$; at intermediate steps, it tracks and propagates the previous masks $\mathrm{Mask}_{i,t-1}$ to $I_t$. The resulting masks are collected as $\mathrm{Mask}_t=\{\mathrm{Mask}_{i,t}\}_{i=1}^{N}$ and forwarded to the Representation Module.

\subsection{Representation Module}

Generalizing across diverse environments requires unifying the underlying state and action spaces. To achieve this, we condition the representation on the embodiment as $\text{Encode}(\text{Obs} \mid \text{Embodiment})$ instead of naively encoding the full observation as $\text{Encode}(\text{Obs})$. 

The module first projects RGB-D observations into a point cloud and transforms it into gripper $i$'s ego-centric frame $\mathcal{G}_i$. Within $\mathcal{G}_i$, we decouple the representation into global and local components. To achieve the most concise encoding possible, we deliberately employ very raw and basic encoding methods for both: (1) a compact global encoding via Artificial Potential Fields (APF) for spatial priors, and (2) fine-grained local Point Cloud (PCD) features of the interaction triplet for precise geometry. To facilitate cross-embodiment transfer, the embodiment is parameterized by gripper dimensions $[length, width, height]$. Formally, the module outputs $( \mathcal{P}^{\text{held}}_{i,t},\, \mathcal{P}^{\text{targ}}_{i,t},\,\mathrm{APF}_{i,t})$ in $\mathcal{G}_i$.

\subsubsection{APF representation for global environment}

Disregarding task-irrelevant obstacles invites collisions, which necessitates a global environment representation. We employ Artificial Potential Fields (APF) as the global encoding, serving as geometry-based heuristics that provide a coarse workspace representation. This formulation benefits from:  (1) natural parameterization by gripper dimensions, (2) inheriting VLM priors for strong OOD generalization, and (3) training-free efficiency. We construct two independent APFs: a \textit{collision APF} for repulsive effects and a \textit{target APF} for attractive effects.

We discretize the point cloud into a voxel grid, assigning a weight $w[p]$ to each occupied voxel center $p$. For the collision APF, let $\mathcal{O}^{\text{self}}_{i,t}$ be voxels of the robot body and held object (from $\mathrm{Mask}_{i,t}$). The weights are:
\begin{equation}
w^{\text{coll}}[p] \triangleq
\begin{cases}
0,  & p \in \mathcal{O}^{\text{self}}_{i,t},\\
-1, & \text{otherwise}.
\end{cases}
\end{equation}
For the target APF, let $\mathcal{O}^{\text{neg}}$ and $\mathcal{O}^{\text{targ}}$ be non-target and target voxels, respectively. The weights are:
\begin{equation}
w^{\text{targ}}[p] \triangleq
\begin{cases}
0,  & p \in \mathcal{O}^{\text{self}}_{i,t},\\
-1, & p \in \mathcal{O}^{\text{neg}},\\
k,  & p \in \mathcal{O}^{\text{targ}}.
\end{cases}
\end{equation}
\begin{equation}
\text{where} \quad k \triangleq \frac{\sum\limits_{p' \in \mathcal{O}^{\text{neg}}} \left| w^{\text{targ}}[p'] \right|}
{\left| \mathcal{O}^{\text{targ}} \right|}.
\end{equation}
This normalization yields zero total weight, stabilizing policy learning.

\begin{figure}[t]
  \centering
  \includegraphics[width=\linewidth]{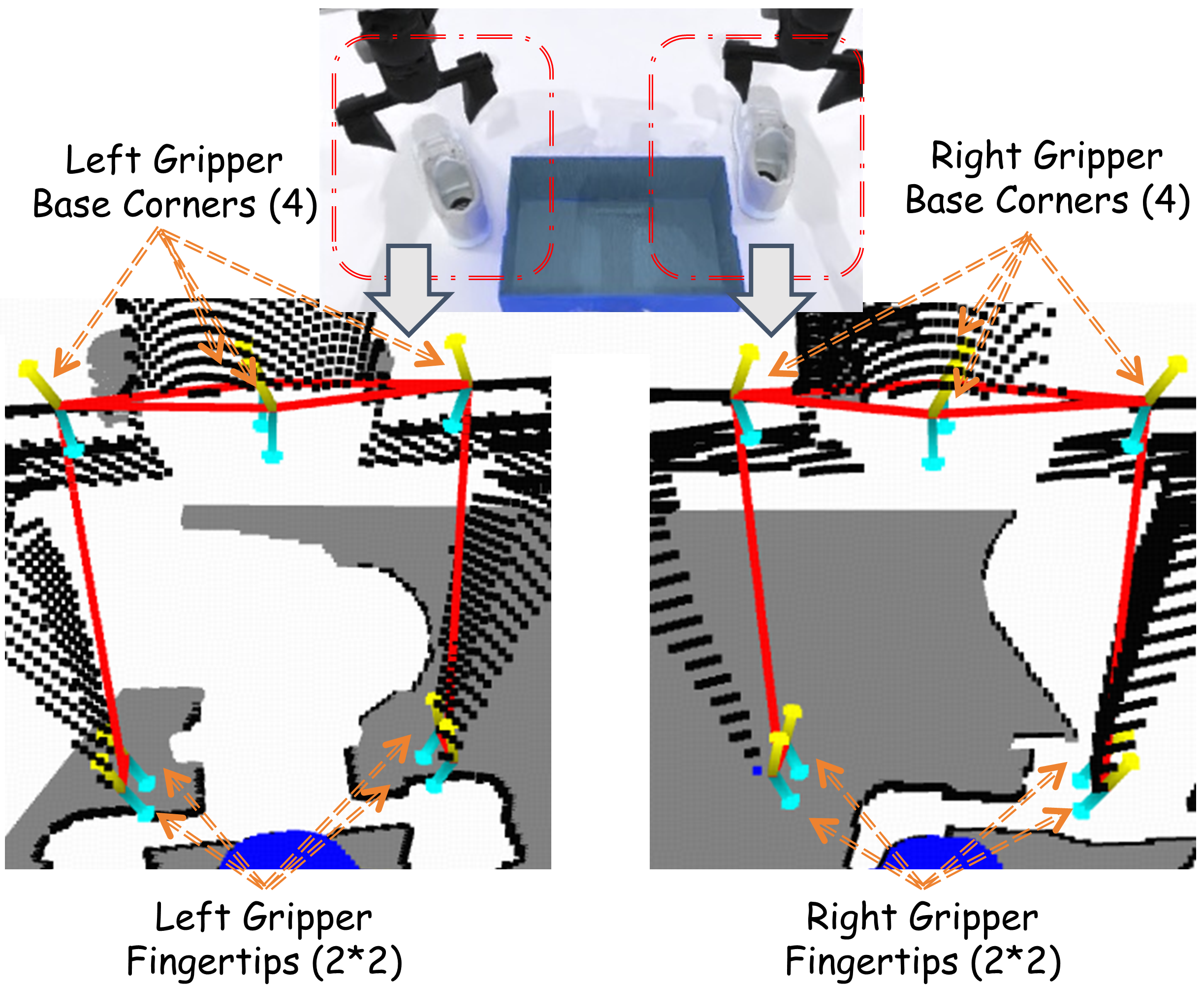}
  \caption{\textbf{Artificial Potential Field (APF) visualization.} The active gripper (red) is attracted by a target APF (cyan) and repelled by a collision APF (yellow).}
  \label{fig:apfs}
\end{figure}

To apply APFs to universal two-finger grippers, we define eight keypoints parameterized by $[length, width, height]$ in $\mathcal{G}_i$ (Fig.~\ref{fig:apfs}): four base corners $[-length, \pm width/2, 0], [-length, 0, \pm height/2]$ and four fingertips $[0, \pm width/2, \pm height/4]$, denoted as $\{q_j\}_{j=1}^8$.
For each keypoint $q$, the APF vector is computed independently for both fields:
\begin{equation}
\mathbf{F}(q) = \sum_{p \in \mathcal{P}_{i,t}} \frac{w[p]}{\|p - q\|^2} \cdot \frac{p - q}{\|p - q\|}.
\end{equation}
We decompose $\mathbf{F}(q)$ into a direction unit vector and a log-scaled magnitude to mitigate imbalances, forming a 4D vector per keypoint. With 8 keypoints and 2 APFs, we obtain a $(16,4)$ tensor as $\mathrm{APF}_{i,t}$. The robustness to varying $[length, width, height]$ values is validated in our ablation studies.

\subsubsection{PCD representation for local geometry}

While APFs encode the global context, contact-rich manipulation requires fine-grained local geometry. Using $\mathrm{Mask}_{i,t}$, we segment the point cloud into $\mathcal{P}^{\text{held}}_{i,t}$ and $\mathcal{P}^{\text{targ}}_{i,t}$. To balance efficiency and fidelity, we downsample them via a two-stage strategy: Farthest Point Sampling (FPS) followed by weighted random sampling (probabilities $\propto$ inverse squared distance to the gripper origin) to prioritize near-gripper points. Each segmented cloud is downsampled to 256 points for policy learning.

\begin{algorithm}[t]
  \caption{Perception Module}
  \label{alg:perception}
  \footnotesize
  \begin{algorithmic}[1]
    \Require $I_t,\ \mathcal{L},\ N,\ \delta$, previous $(I_{t-1},\ \mathcal{C}_{t-1},\ \mathrm{Mask}_{t-1})$     \Ensure current task context $\mathcal{C}_t$ and $\mathrm{Mask}_t$     \If{$t \bmod \delta = 0$} \Comment{as refresh steps}
      \State $\mathcal{C}_t\gets\emptyset$       \For{$i=1$ to $N$}
        \State $\mathcal{C}_{i,t} \leftarrow$ Query VLM with gripper-$i$ prompt
        \State $\mathcal{C}_t\gets \mathcal{C}_t\cup\{\mathcal{C}_{i,t}\}$         \State $\mathrm{Mask}_{i,t}\gets$ \textbf{Seg}($I_t$, boxes from $\mathcal{C}_{i,t}$)
      \EndFor
    \Else \Comment{at intermediate steps}
      \State $\mathcal{C}_t \gets \mathcal{C}_{t-1}$       \For{$i=1$ to $N$}
        \State $\mathrm{Mask}_{i,t}\gets$ \textbf{Track}($I_{t-1}$, $\mathrm{Mask}_{i,t-1}$, $I_t$)
      \EndFor
    \EndIf
    \State $\mathrm{Mask}_t \gets \{\mathrm{Mask}_{i,t}\}_{i=1}^{N}$     \State $(I_{t-1},\ \mathcal{C}_{t-1},\ \mathrm{Mask}_{t-1})\gets(I_t,\ \mathcal{C}_t,\ \mathrm{Mask}_t)$     \State \Return $\mathcal{C}_t$ and $\mathrm{Mask}_t$   \end{algorithmic}
\end{algorithm}

\subsection{Action Module}
To enable cross-embodiment skill transfer, the Action Module decouples global task semantics from embodiment-specific perceptions. It employs a Flow-Matching Transformer to predict a horizon-$H$ 7-DoF action chunk $\mathbf{A}_{i,t}$ at perception step $t$. Specifically, the policy input consists of a global subtask text $\mathcal{C}_t$ and per-effector perception zones encoding $\text{Encode}(\text{Obs} | \text{Embodiment}_i)$. For gripper $i$, its zone incorporates the interaction-centric representation $(\mathcal{P}^{\text{held}}_{i,t}, \mathcal{P}^{\text{targ}}_{i,t}, \mathrm{APF}_{i,t})$ and proprioceptive state $\xi_{i,t}$.

This per-effector formulation enables seamless cross-embodiment deployment. In dual-arm settings, the model conditions on the global text and two independent gripper zones. To deploy this model on a single-arm robot, we simply fix the inactive arm's perception zone to its initial state while aligning the active arm's zone with the current gripper.

Internally, we encode $\mathcal{C}_t$ with a frozen pretrained text encoder and a trainable MLP adapter. Each gripper zone is represented by 4 tokens: the held and target object point clouds ($\mathcal{P}^{\text{held}}_{i,t}$, $\mathcal{P}^{\text{targ}}_{i,t}$) encoded via a PointNet~\cite{pointnet}, the gripper's $\mathrm{APF}_{i,t}$ encoded via a Transformer over gripper-keypoint tokens, and its world-frame pose $\xi_{i,t}$ encoded via an MLP. Encoder weights are shared across grippers. The resulting global text tokens and per-effector modality tokens are fused with a noisy action-token sequence, and processed by an 8-layer causal Transformer denoiser (512 hidden dims, $\sim$30M parameters) trained with a flow-matching objective to yield smooth actions.

Training uses demonstrations from simulation and real-world teleoperation. Each trajectory is processed by the Perception and Representation Modules to produce $\mathcal{C}_t$ and $( \mathcal{P}^{\text{held}}_{i,t}, \mathcal{P}^{\text{targ}}_{i,t}, \mathrm{APF}_{i,t})$, paired with demonstrated actions to supervise the flow-matching policy.

\subsection{Deployment Efficiency}
\label{sec:efficiency}

While Perception Module currently relies on the closed-source GLM-4.6V API with variable latency, we assess the system's deployable speed based on the standard inference time of locally deployed open-source VLMs (200--500ms latency). In this setting, non-VLM components execute in 10ms. Thanks to tracking and action chunking, a single VLM call triggers 5 SAM calls, yielding 100 actions in total. Executed at 20Hz, these actions span 5s, effectively amortizing the VLM latency and enabling real-time deployment.

%% file: latex/4.experiment.tex
\begin{figure*}
  \includegraphics[width=\textwidth]{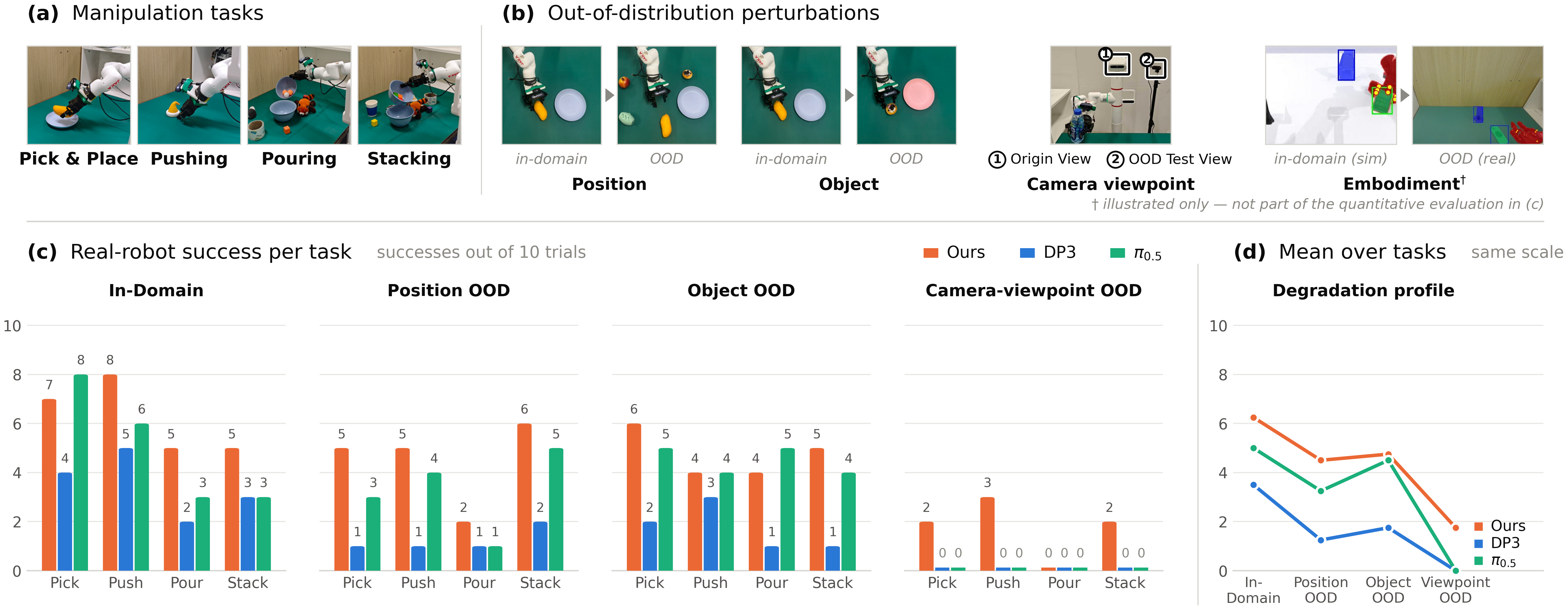} 
  \caption{\textbf{Real-world experiments.} We collect data on four tasks and evaluate the model under in-domain and three different out-of-distribution (OOD) settings (Position/Object/Camera OOD). Additionally, we directly deploy the simulation-trained model onto a physical robot with a heterogeneous embodiment relative to the original platform (Embodiment OOD).}
 \label{fig:exp} 
\end{figure*}

\section{Experiments}
\label{sec:Experiment}

We evaluate our method across three dimensions: core performance, robustness, and component contributions. First, we benchmark against state-of-the-art baselines in both simulation and physical environments. Second, we rigorously test robustness under diverse out-of-distribution (OOD) scenarios, including position, object, and camera-viewpoint shifts. Notably, we tackle highly challenging zero-shot cross-embodiment transfer, directly deploying a model trained exclusively on a dual-arm setup to a single-arm platform without post-training. Finally, we isolate our design choices and investigate the source of OOD generalization through comprehensive ablation studies.




\begin{table}[t]
    \centering
    \caption{\textbf{Simulation Experiment.} Comparison on 25 tasks from RoboTwin2.0 benchmark. We report success rates (\%) under Clean and Random settings. \textbf{Bold} indicates the best and \underline{underline} indicates the second best.}    \label{tab:main_comparison}
    \setlength{\tabcolsep}{6pt}
    \renewcommand{\arraystretch}{1.1}
    \begin{tabular}{lcc}
        \toprule
        \textbf{Method} & \textbf{Clean} & \textbf{Random} \\
        \midrule
        \multicolumn{3}{l}{\textit{Small Models (Trained from Scratch)}} \\
        Ours & 66.3 & \textbf{46.1} \\
        ACT & 42.4 & 3.2 \\
        DP3-G & 49.0 & 10.0 \\
        DP3-W & \underline{73.1} & 7.5 \\
        \midrule
        \multicolumn{3}{l}{\textit{Large Models (Fine-tuned)}} \\
        RDT & 47.0 & 19.3 \\
        $\pi_0$ & 61.4 & 24.4 \\
        UP-VLA & 66.5 & 24.8 \\
        BagelVLA & \textbf{85.5} & \underline{28.4} \\
        \bottomrule
    \end{tabular}
\end{table}

\subsection{Simulation Experiments}

\vpara{Experiment Settings.}
We evaluate on the RoboTwin2.0~\cite{robotwin2} benchmark. For each task, we train on 50 \textit{Clean} demonstration trajectories and test on 100 trajectories in both \textit{Clean} and \textit{Randomized (Rand.)} settings, where the latter induces OOD shifts via scene randomization. 

We compare against Small Models trained from scratch (ACT~\cite{act}, DP3~\cite{dp3} in world/end-effector frames, denoted as DP3-W and DP3-G) and Large Models fine-tuned from pre-trained checkpoints ($\pi_0$~\cite{pi0}, RDT~\cite{liu2025rdt}, UP-VLA~\cite{zhang2025up}, BagelVLA~\cite{hu2026bagelvla}).

\vpara{Task Selection.}
We select a subset of 25 tasks from RoboTwin2.0, excluding tasks incompatible with our system's assumptions of relying on a VLM to extract priors and track states via a single fixed-view RGB-D stream. Excluded tasks fall into three categories:
\begin{itemize}
    \item \textbf{Ambiguous state prediction.} Progress cannot be reliably inferred when positional differences lack visual features (e.g., \texttt{place\_object\_scale} requires placing at unmarked locations).
    \item \textbf{Insufficient viewpoint coverage.} Tasks requiring precise alignment or occluded interactions (e.g., \texttt{beat\_block\_hammer}) necessitate wrist/multi-view perception unavailable in our setup.
    \item \textbf{Low-contrast localization.} Objects blending into the background cause VLM/SAM parsing failures (e.g., the light-colored switch in \texttt{turn\_switch} merges with a white background).
\end{itemize}

\vpara{Results.}
Table~\ref{tab:main_comparison} reports success rates averaged over 25 tasks. Our method achieves the highest \textit{Randomized} success rate (\textbf{46.1\%}) among all methods. While DP3-W attains a higher \textit{Clean} score (73.1\%), its performance drops sharply to 7.5\% under \textit{Randomized}, revealing a severe lack of robustness. Although large models exhibit strong \textit{Clean} performance (e.g., BagelVLA at 85.5\%), they suffer substantial degradation under \textit{Rand.}. Our Clean performance is on par with UP-VLA (66.3\% vs. 66.5\%) and significantly outperforms $\pi_0$ and RDT, despite being a much smaller hierarchical model trained entirely from scratch. This markedly higher retention rate from \textit{Clean} to \textit{Randomized} underscores our method's exceptional robustness against induced distribution shifts.


\begin{figure*}
  \includegraphics[width=\textwidth]{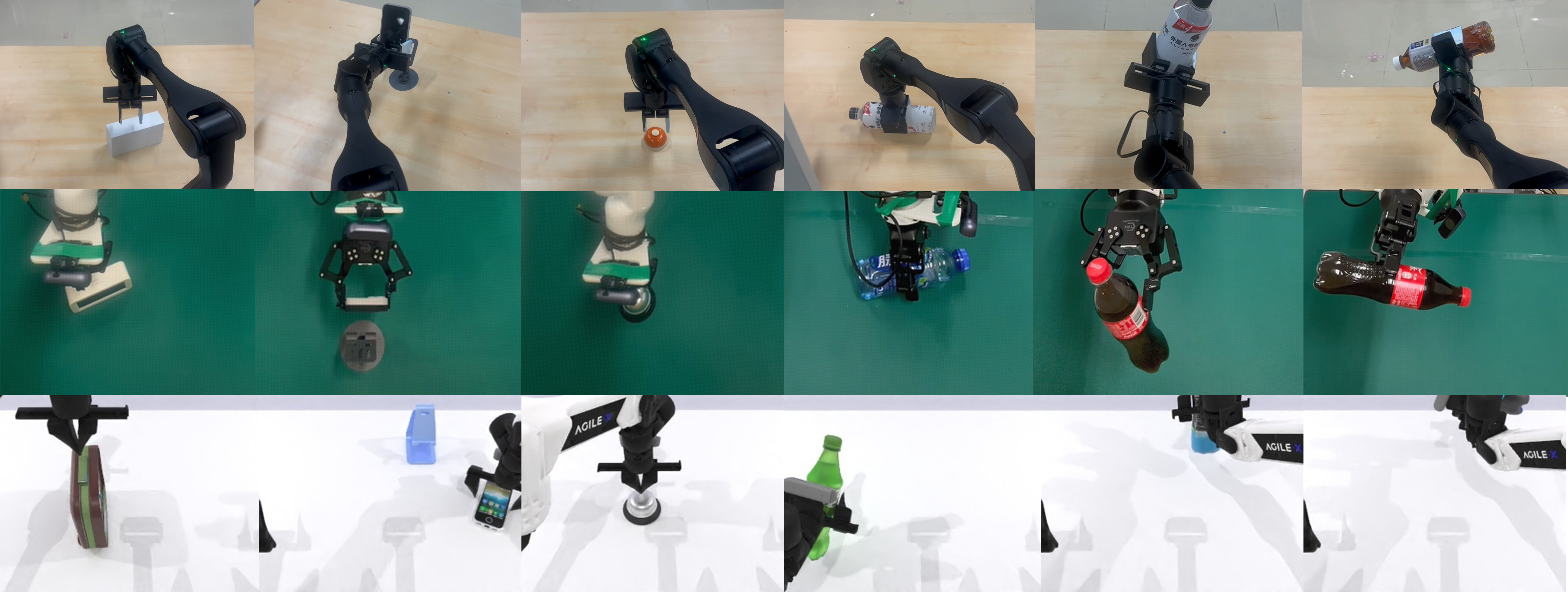} 
\caption{We selected six tasks from RoboTwin2.0: \textit{Click Alarmclock}, \textit{Place Phone Stand}, \textit{Click Bell}, \textit{Adjust Bottle}, \textit{Shake Bottle}, \textit{Shake Bottle Horizontally}, and deployed the policy on our robots. Rows from top to bottom: real-world deployment on DOS-W1, on Jaka, and the simulation.}
 \label{fig:sim2real} 
\end{figure*}

\subsection{Real-World Experiments} 

\vpara{Experiment Settings.} As shown in Fig. \ref{fig:exp}, we evaluate our system on a real Jaka robotic arm to assess robustness under out-of-distribution (OOD) settings. We consider two evaluation protocols:

(1) \textbf{Position, object, and camera-viewpoint OOD.}
We design four real-world manipulation tasks: \textit{pick-and-place}, \textit{hat pushing}, \textit{bowl stacking}, and \textit{pouring}. The three OOD conditions are: (i) \textbf{Position OOD} (adding distractors), (ii) \textbf{Object OOD} (replacing with unseen instances), and (iii) \textbf{Camera-viewpoint OOD} (moving the camera by $35$\,cm and $45^\circ$). 
We collect 50 real-robot demonstration trajectories per task and compare against DP3~\cite{dp3} and $\pi_{0.5}$~\cite{pi05}.




\begin{table}[t]
    \centering
    \caption{\textbf{Embodiment OOD results.} \textbf{Sim} denotes success rates in RoboTwin2.0. \textbf{DOS-W1} and \textbf{JAKA} show real-world success counts. \textbf{Near-view} involves minor camera shifts ($\sim$10cm/25$^\circ$), while \textbf{Novel-view} uses a substantially different perspective.}
    \label{tab:sim2real}
    \resizebox{\columnwidth}{!}{%
    \begin{tabular}{lcccccc}
        \toprule
        & \shortstack{\textit{Adjust}\\\textit{Bottle}} 
        & \shortstack{\textit{Shake}\\\textit{Bottle}} 
        & \shortstack{\textit{Shake}\\\textit{Horiz.}} 
        & \shortstack{\textit{Click}\\\textit{Alarm.}} 
        & \shortstack{\textit{Click}\\\textit{Bell}} 
        & \shortstack{\textit{Place}\\\textit{Phone}} \\
        \midrule
        \textbf{Sim} & 96\% & 97\% & 92\% & 78\% & 80\% & 32\% \\
        \textbf{DOS-W1} & 18/30 & 16/30 & 15/30 & 15/30 & 13/30 & 6/30 \\
        \textbf{JAKA (near-view)} & 14/30 & 15/30 & 14/30 & 11/30 & 9/30 & 5/30 \\
        \textbf{JAKA (novel-view)} & 8/30 & 9/30 & 7/30 & 6/30 & 5/30 & 2/30 \\
        \bottomrule
    \end{tabular}%
    }
\end{table}

(2) \textbf{Embodiment OOD (Cross-Heterogeneous-Robot Zero-Shot Transfer).}
As shown in Fig. \ref{fig:sim2real}, we train purely on RoboTwin2.0 simulated trajectories with the ALOHA-AgileX embodiment, then deploy the same policy zero-shot on platforms with distinct robots: Jaka (more distal) and DOS-W1 (closer to the original). Leveraging per-effector perception, we align the target gripper with the active right arm's zone while fixing the inactive left arm's zone. For Jaka, we evaluate under \textbf{near-view} ($\sim$10\,cm/25$^\circ$ shift) and \textbf{novel-view} configurations (Fig. \ref{fig:exp}). For DOS-W1, the unadjusted camera shares similar x-y coordinates to the simulation but is raised 40\,cm with a steeper downward tilt. This setup simultaneously compounds the sim-to-real gap, dual-arm to single-arm morphological shifts, camera viewpoint shifts, and object variations. To the best of our knowledge, ours is the first IL approach to bridge this gap.



\vpara{Results.}
Fig~\ref{fig:exp} and Table~\ref{tab:sim2real} evaluate robustness under explicit OOD shifts (Position/Object/Camera-viewpoint) and the most challenging Embodiment OOD protocol.

\textbf{Real-world OOD robustness (Fig~\ref{fig:exp}).}
Our method outperforms DP3 and is highly competitive with $\pi_{0.5}$ in the In-Domain setting. Under \textbf{Position OOD} and \textbf{Object OOD}, our method maintains higher success on every task, achieving 5--6/10 on \textit{pick-and-place} and \textit{stacking}, and 2--5/10 on \textit{pushing} and \textit{pouring} (vs.\ 1--5/10 for both baselines). Under \textbf{Camera-viewpoint OOD}, both baselines drop to 0/10 on all tasks, whereas our method retains non-zero success on \textit{pick-and-place} (2/10), \textit{block pushing} (3/10), and \textit{bowl stacking} (2/10); \textit{pouring} remains the hardest and reaches 0/10 for all approaches.

\textbf{Embodiment OOD (Table~\ref{tab:sim2real}).} 
We omit baselines as no prior manipulation IL system achieves zero-shot transfer across heterogeneous embodiments.

Despite near-perfect simulated success ($>$90\% on \textit{Shake Bottle} and \textit{Shake Horiz.}), zero-shot deployment on physical robots incurs significant degradation from the compounding challenges of the sim-to-real gap, dual-to-single-arm transfer, object variations, and viewpoint shifts. 

Our method retains $\sim$50\% of its simulation performance on dynamic tasks for both \textbf{DOS-W1} and \textbf{Jaka (near-view)} (e.g., \textit{Shake Bottle} 16/30 and 15/30, \textit{Shake Horiz.} 15/30 and 14/30). DOS-W1 generally achieves slightly higher success, consistent with its closer workspace proximity to the original robot. As expected, precise placement remains challenging (\textit{Place Phone} 6/30 and 5/30). 
Performance predictably drops under the extreme perspective shift of \textbf{Jaka (novel-view)} (e.g., \textit{Shake Bottle} degrades to 9/30), highlighting the method's inherent robustness to morphological embodiment shifts while remaining sensitive to severe viewpoint variations.



\begin{table}[t]
    \centering
    \caption{Ablation studies on Robotwin2.0. \textbf{I:} Component validation. \textbf{II:} Generalization source analysis. }
    \label{tab:ablation}
    \setlength{\tabcolsep}{6pt}
    \renewcommand{\arraystretch}{1.1}
    \begin{tabular}{lcc}
        \toprule
        \textbf{Method} & \textbf{Clean} & \textbf{Random} \\
        \midrule
        \multicolumn{3}{l}{\textit{Ablation I: Component Validation}} \\
        Ours (Full) & 66.3 & 46.1 \\
        w/o CM & 54.2 & 36.8 \\
        w/o APF & 61.0 & 35.1 \\
        w/o PCD & 54.9 & 39.2 \\
        Mismatched Dim & 62.8 & 43.6 \\
        \midrule
        \multicolumn{3}{l}{\textit{Ablation II: Generalization Source}} \\
        Action-Module (PCD) & 39.2 & 4.2 \\
        + seg (all) & 54.4 & 6.6 \\
        + seg (instruction) & 55.8 & 24.0 \\
        + seg (VLM subtask) & 61.0 & 35.1 \\
        \bottomrule
    \end{tabular}
\end{table}

\subsection{Ablation Study}

We conduct two lines of ablation studies on 25 tasks: \textbf{I) Component Validation}, which evaluates core architectural modules, and \textbf{II) Generalization Source}, which probes the origin of OOD robustness through progressive augmentation. Table~\ref{tab:ablation} reports the average success rates under \textit{Clean} and \textit{Randomized (Rand.)} settings.


\vpara{Ablation I: Component Validation.}
This ablation line isolates the core components of our action module---Artificial Potential Field (APF) encoding, point cloud geometric (PCD) encoding, and causal masking (CM)---alongside the impact of embodiment parameterization. To validate the necessity of accurate gripper dimensions, we introduce a ``Mismatched Dim'' variant. The default parameters [0.095 m, 0.090 m, 0.030 m] represent the exact simulator gripper, whereas the ablated variant uses the larger physical dimensions [0.185 m, 0.110 m, 0.050 m] from our real-world setup.

The macroscopic results in Table~\ref{tab:ablation} highlight the complementary roles of these components. Removing PCD significantly degrades \textit{Clean} performance (66.3\% $\rightarrow$ 54.9\%) due to the loss of fine-grained local geometry, whereas removing the lossy APF encoding severely harms \textit{Randomized} robustness (46.1\% $\rightarrow$ 35.1\%), confirming its value in providing a stable global context. Furthermore, ablating CM not only drops overall success rates (to 54.2\% and 36.8\%) but also introduces temporal incoherence into the predicted action chunks. This jitteriness is partially masked by binary simulation metrics but would be highly detrimental to stable real-world deployment.

A per-task breakdown reveals that performance degradation is strictly task-specific: removing CM disproportionately impacts multi-step sequences, lacking PCD leads to failures in fine-geometry tasks, and missing APF primarily degrades performance in cluttered scenes. Similarly, \textit{Mismatched Dim} yields moderate overall drops but deteriorates significantly in tasks requiring precise spatial alignment, underscoring that accurate embodiment parameterization is critical for fine-grained manipulation.

\vpara{Ablation II: Generalization Source.}
Beyond component validation, we investigate the primary drivers of OOD generalization. Starting from a pure \textit{Action-Module (PCD)} baseline that processes the entire raw scene point cloud, we progressively introduce visual abstraction: generic object segmentation (\textit{+ seg (all)}); instruction-relevant semantic segmentation via SAM3~\cite{carion2026sam} (\textit{+ seg (instruction)}); and the full VLM-based sub-task semantic triplet (\textit{+ seg (VLM subtask)}), which isolates the held and target objects per sub-task.

The results isolate the contribution of high-level semantics. The raw point cloud baseline yields poor performance on both \textit{Clean} (39.2\%) and \textit{Random} (4.2\%). While generic segmentation (\textit{+ seg (all)}) filters background distractors and improves \textit{Clean} to 54.4\%, OOD robustness remains severely limited (\textit{Random}: 6.6\%). Crucially, introducing instruction-relevant and sub-task-level semantic segmentation significantly enhances OOD generalization, boosting the \textit{Random} success rate to 24.0\% and 35.1\%. This demonstrates that robust OOD generalization stems not from naive background filtering, but from grounding fine-grained actions in a structured subtask progression.

%% file: latex/5.conclusion.tex

\section{Conclusion, Limitations, and Future Work}

In this work, we propose an interaction-centric framework that establishes a unified cross-domain representation for two-finger manipulation. By hybridizing geometric parameterization with learning-based policies, we decouple embodiment from visual states via an interaction triplet. This yields a robust in-domain imitation learning system with strong OOD generalization, achieving zero-shot sim-to-real transfer to unseen heterogeneous robots despite object and viewpoint variations.

Despite these advances, limitations remain: 1) VLM reliance restricts input to single-view RGB and limits complex task decomposition; 2) localization and segmentation biases from VLMs and SAM compound; 3) viewpoint changes require manual camera re-calibration—a necessary trade-off for zero-shot viewpoint transferability; and 4) cross-embodiment deployment demands strict workspace matching, constraining feasible tasks and occasionally incurring inverse kinematics (IK) failures.

Future work will focus on multi-view perception and more robust foundation models. Ultimately, we aim to learn interaction-centric semantics directly from raw world states rather than relying on geometric encoding. Using these as intermediate representations for VLA prediction is key to unlocking broader cross-embodiment transfer and mixed-data pretraining.